\documentclass[lettersize,journal]{IEEEtran}
\usepackage{amsmath,amsfonts}
\usepackage[ruled, linesnumbered]{algorithm2e}
\usepackage{array}
\usepackage[caption=false,font=normalsize,labelfont=sf,textfont=sf]{subfig}
\usepackage{textcomp}
\usepackage{stfloats}
\usepackage{url}
\usepackage{verbatim}
\usepackage{graphicx}
\usepackage{cite}
\usepackage{booktabs}
\usepackage{makecell}
\usepackage{xcolor}  
\usepackage{hyperref}
\usepackage{color}
\usepackage{threeparttable} 
\begin{document}

\title{DAG-ACFL: Asynchronous Clustered Federated Learning based on DAG-DLT}
\author{Xiaofeng Xue, Haokun Mao  and  Qiong Li
	\thanks{The authors are with the Information Countermeasure Technique Institute, School of Cyberspace Science, Faculty of Computing, Harbin Institute of Technology, Harbin 150001, China (e-mail: xfxue@hit.edu.cn; hkmao@hit.edu.cn; qiongli@hit.edu.cn).}}
% The paper headers
\markboth{}%
{Shell \MakeLowercase{\it{et al.}}: A Sample Article Using IEEEtran.cls for IEEE Journals}

% \IEEEpubid{0000--0000/00\$00.00~\copyright~2021 IEEE}
% Remember, if you use this you must call \IEEEpubidadjcol in the second
% column for its text to clear the IEEEpubid mark.

\maketitle

\begin{abstract}
Federated learning (FL) aims to collaboratively train a global model while ensuring client data privacy. However, FL faces challenges from the non-IID data distribution among clients. Clustered FL (CFL) has emerged as a promising solution, but most existing CFL frameworks adopt synchronous frameworks lacking asynchrony. An asynchronous CFL framework called SDAGFL based on directed acyclic graph distributed ledger techniques (DAG-DLT) was proposed, but its complete decentralization leads to high communication and storage costs. We propose DAG-ACFL, an asynchronous clustered FL framework based on directed acyclic graph distributed ledger techniques (DAG-DLT). We first detail the components of DAG-ACFL. A tip selection algorithm based on the cosine similarity of model parameters is then designed to aggregate models from clients with similar distributions. An adaptive tip selection algorithm leveraging change-point detection dynamically determines the number of selected tips. We evaluate the clustering and training performance of DAG-ACFL on multiple datasets and analyze its communication and storage costs. Experiments show the superiority of DAG-ACFL in asynchronous clustered FL. By combining DAG-DLT with clustered FL, DAG-ACFL realizes robust, decentralized and private model training with efficient performance.
\end{abstract}

\begin{IEEEkeywords}
Federated learning, clustering, asynchronous, directed acyclic graph, distributed ledger technology.
\end{IEEEkeywords}

\section{Introduction}
\IEEEPARstart{T}{he} increasing popularity of the Internet of Things (IoT) and smartphones has significantly transformed people's lifestyles and work environments, providing enhanced connectivity and convenience. Current research indicates that there are more than 14.3 billion active IoT endpoints \cite{IoT2023} and approximately 6.84 billion smartphones \cite{sp2023} globally. These devices always have advanced sensors, computing, and communication capabilities, enabling them to perform various tasks and collect vast amounts of data to train machine-learning models. However, the data collected by these devices is typically distributed across different locations and owned by different users. Limited communication resources present challenges to collecting these data to a central server. Moreover, the data collected by these devices often contain sensitive personal information, causing users to be hesitant to share their data with others due to privacy concerns. Consequently, there are significant challenges and limitations in centrally training machine learning models using such distributed and privacy-sensitive data.

Federated learning\cite{konecny2016,mcmahan2017,yang2019,bonawitz2019} was proposed to address the abovementioned problems.  In FL, data is stored locally on edge devices such as smartphones and IoT devices, referred to as clients. FL realizes model training by enabling model training on these clients and aggregating these trained model parameters on the central server. Thus, FL can effectively protect the data privacy of user data. Due to its ability to train machine learning models using distributed data, FL has garnered considerable attention and is widely recognized as a promising solution.

However, the inability of FL's server to access the actual distribution of user data makes it vulnerable to the challenge arising from the statistical heterogeneity of data. This non-independent and identically distributed (non-iid) data distribution can lead to a decrease in model accuracy. Zhao et al. showed that there is a 55\%  degradation on a model trained with convolutional neural networks (CNN) on the highly skewed CIFAR-10 dataset\cite{zhao2022}. Meanwhile, this non-iid data distribution can also make the convergence of the global model slow down or even fail\cite{sattler2020}. The non-iid data distribution can be categorized into client-wise non-iid and clustered-wise non-iid data distribution according to the research of ma et.al\cite{ma2022}. The distribution where the labels vary significantly from clients is called client-wise non-iid data distribution, and the distribution where the label distribution varies significantly from the different clusters is called clustered-wise non-iid data distribution. The clustered FL framework\cite{sattler2021,duan2021,ghosh2022a,long2023} was proposed to address the more prevalent clustered-wise non-iid data distribution in practical applications. The clustered FL frameworks always perform better on the clustered-wise non-iid data distribution than the vanilla FL framework\cite{mcmahan2017}. 

These clustered FL frameworks adopt a synchronization framework similar to Vanilla FL, which is challenged in handling network environment heterogeneity and clients' resource is limited  \cite{xu2021, li2020b, laguel2020}, which is a common scenario in an IoT or edge device environment. Moreover, it is vulnerable to single points of failure associated with centralized structures \cite{wang2021c, hou2021}, which can result in training process interruptions, model parameter leaks, and privacy security concerns. 

A decentralized clustered FL framework using Distributed Ledger Techniques (DLT) based on Directed Acyclic Graphs (DAG) \cite{popov2015}, called SDAFGL\cite{beilharz2021}, addresses the abovementioned challenge by using the advantages of the DAG-DLT. SDAFGL handles clustered-wise non-iid datasets using an accuracy-biased random walk tip selection algorithm. The decentralized nature of DAG-DLT allows for asynchronous FL, which can effectively address the challenges posed by network heterogeneity and resource limitations. Additionally, the DAG-DLT structure ensures that the system is not vulnerable to single points of failure. However, the fully decentralized nature of SDAFGL complicates communication among clients and incurs high communication costs. Additionally, the requirement to store the DAG ledger containing all model parameters at each client consumes significant client storage resources. The high communication and storage costs hinder its deployment on resource-constrained edge devices.

Thus, we present a novel approach called DAG-ACFL, a DAG-based asynchronous clustered federated learning framework to address the abovementioned challenges. It comprises two layers: the client layer and the server layer. Additionally, we propose a similarity of model parameter-based tip selection algorithm to determine the most appropriate tip of the DAG ledger. The main contributions of this paper are as follows:
\begin{enumerate} 
    \item We design an asynchronous bi-layer clustered FL framework using DAG-DLT specifically for handling cluster-wise non-IID data distributions in edge devices. It realizes client-oriented asynchronous training and reduces costs compared to SDAFGL. We analyze resource consumption theoretically. 
    \item We propose a tip selection algorithm utilizing model parameter similarity between clients and tips of the DAG ledger to identify tips from the same cluster. The server aggregates parameters from tips with high similarity to the client's new model. 
    \item An adaptive tip selection algorithm is proposed using change point detection to dynamically determine the number of tips. It can select all relevant tips without predefining cluster numbers.
    \item We evaluate the performance of DAG-ACFL on different datasets using two type settings, validating its efficacy over other clustered FL frameworks.
\end{enumerate}

The rest of this paper is organized as follows. Section \ref{sec:relatedwork} reviews related works on FL, clustered FL and DAG-based FL. Section \ref{sec:framework} presents the motivation and design of DAG-ACFL, including the tip selection algorithms. Section \ref{sec:evaluation} shows experimental results and analysis. Section \ref{sec:conclusion} concludes this paper and discusses future work.

\section{Background and Related Work}
\label{sec:relatedwork}
\subsection{Federated Learning}
This section introduces the fundamental concepts and workflow of federated learning (FL). A typical FL framework comprises a central server that contains the global model and multiple participating clients. The objective of FL is to collaboratively train a global model using the data distributed across these participating clients. We can express the global optimization objective of the FL framework as follows:
\begin{equation}
    \min _\omega F(\omega)=\sum_{k=1}^N p_k F_k(\omega)
\end{equation}
Where $\omega$ represents the model parameters, $N$ is the number of clients, and $p_k$ denotes the weight of the $k$th client. The weights $p_k$ are non-negative and satisfy the condition $\sum_{k=1}^N p_k=1$. $F_k(\omega)$ represents the local optimization objective for each client. It calculates the prediction of the local sample set $(x_i, y_i)$ using the model parameters $\omega$ and measures the difference between the prediction result $f(x_i, \omega)$ and the corresponding actual label $y_i$.

Assuming that the $k$th client has $n_k$ samples, we can utilize the average loss function $L(\cdot)$ to compute its local optimization objective $F_k(\omega)$ as follows:
\begin{equation}
    F_k(\omega)=\frac{1}{n_k} \sum_{i=1}^{n_k} L(f(x_i, \omega),y_i)
\end{equation}

The FedAvg algorithm \cite{mcmahan2017} is widely recognized as one of the most prominent algorithms in FL. Its core concept revolves around optimizing the global optimization objective $F_\omega$ by iteratively exchanging model parameters between the server and the clients. In the $t$th iteration, The server first sends the current global model parameter $\omega^{t-1}$ to the selected clients. Then, each client utilizes its local training dataset to update the model parameter $\omega^{t-1}$ independently. This update is performed through multiple local iterations, typically employing optimization algorithms such as stochastic gradient descent (SGD). The process in client $k$ can be expressed as follows:
\begin{equation}
    \omega^{t}_k = \omega^{t-1}_k - \eta \nabla F_k(\omega^{t-1}_k)
\end{equation}
where,$\eta$ is learning rate, $\nabla F_k(\omega^t)$ is the gradient of the $F_k(\omega^t)$. 

After completing the local updates, the clients send their updated model parameters $\omega^{t}_i$ back to the server. The server aggregates the received model parameters to generate the new global model parameter $\omega^t$. It can be expressed as follows:
\begin{equation}
    \omega^{t} = \sum_{k=1}^N p_k \omega_k^{t+1}
    \label{eq:fedavg}
\end{equation}

The above steps are repeated for multiple iterations to achieve convergence toward an optimized global model. 
% The FedAvg algorithm is shown in Algorithm \ref{alg:fedavg}.

% \begin{algorithm}[h]
%     \caption{Federated Averaging (FedAvg)}
%     \label{alg:fedavg}
%     \KwIn{$N$ clients with local datasets, $D_i$ is the  dataset hold in clients $i$,$E$ is number of local epochs, $\eta$ is learning rate, $\gamma$ is participation rate of clients in each round }
%     \KwOut{Optimized Global Model parameter $\omega$}
%     Initial global model parameter $\omega_0$\;
%     \For{each round $t = 0, 1, 2, \dots$}{
%         $m \leftarrow \max(C \cdot N, 1)$\;
%         $S_t \leftarrow$ (random set of $m$ clients)\;
%         \For{each client $k \in S_t$ {\bf{in parallel}}}{
%             $\omega_{k}^{t} \leftarrow$ ClientLocalTrain($k, \omega^{t}$, $E$, $\eta$)\;
%         }
%         $|D|$ = $\sum_{i\in S_t} |D_i|$\;
%         $\omega^{t+1} \leftarrow \sum_{k=1}^N \frac{|D_i|}{|D|} \omega_k^{t}$\;
%     }
%     \SetKwFunction{ClientLocalTrain}{ClientLocalTrain}
%     \SetKwProg{Function}{Function}{:}{end}
    
%     \Function{\ClientLocalTrain{$k, \omega, E, \eta$}}{
%         \For{$e=1$ \KwTo $E$}{
%             $\omega \leftarrow \omega - \eta \nabla F_k(\omega)$\;
%         }
%         \Return{$\omega$}\;
%     }
% \end{algorithm}
% \label{sec:Motivation}
\subsection{Clustered Federated Learning}
In Vanilla FL, the cooperative learning of the global model is achieved through the exchange of model parameters between the server and clients. However, it is worth noting that the performance of the global model trained by Vanilla FL may only be optimal for some clients, especially when the data distribution among clients is non-iid. Recent research by Ma et al. \cite{ma2022} has classified non-IID data distributions into client-wise non-IID and clustered-wise non-IID, with the latter being more prevalent in real-world applications. To solve the cluster-wise non-IID dataset challenge, clustered FL has emerged as an active area of research, offering advantages specifically for the clustered-wise non-IID data distribution.

The core idea of clustered FL is to divide participating clients into multiple clusters based on specific similarity measures and train the global model within each cluster. Sattler et al. proposed a multitask federated learning framework called Clustered Federated Learning (CFL) \cite{sattler2021}. CFL gradually assigns clients to suitable clusters using the cosine similarity of gradient update parameters. Another clustered FL approach, FeSEM, was proposed by Long et al. \cite{long2023}. It clusters clients based on the distance between their local model parameters and the clustered global model parameters stored on the server. FlexCFL, proposed by Duan et al. \cite{duan2021}, uses the Euclidean distance of Decomposed Cosine similarity (EDC) algorithm to cluster clients into a predefined number of clusters. FL+HC \cite{briggs2020} utilizes a hierarchical cluster algorithm to cluster clients. In contrast to the abovementioned methods, IFCA proposed by Ghosh et al. \cite{ghosh2022a} assigns clients to the most suitable cluster by evaluating the loss of different models within different clusters. Jamali-Rad et al. proposed Federated Learning with Taskonomy (FLT) \cite{jamali-rad2022}, where the server provides each client with a pre-trained encoder to extract potential features from its dataset. The server learns the correlation of training tasks among clients through manifold learning and clusters the clients into different disjoint clusters using a hierarchical clustering algorithm.

\subsection{Federated learning with DAG-DLT}
Distributed Ledger Technology(DLT) is a decentralized distributed database that can share data among different network nodes without the need for a central manager. Its main features and advantages include decentralization, auditability and tamper-proof. Blockchain\cite{wright2008,androulaki2018} and DAG-DLT\cite{popov2015} are the two popular DLTs. Compared to the blockchain, DAG-DLT has the following advantages: (1) DAG-DLT adopts a voting consensus mechanism that is friendly to devices with limited resources, while the blockchain adopts a proof of X consensus mechanism that requires a large amount of computing resources. (2) DAG-DLT realizes an asynchronous bookkeeping system, while the blockchain is a synchronous bookkeeping system; so the DAG-DLT is more suitable for realizing an asynchronous FL.

Compared with blockchain-based FL\cite{urrehman2020, nguyen2021, li2022b, chen2022b, issa2023}, it offers several advantages. First of all, the natural asynchronous nature of DAG-DLT allows it to naturally implement asynchronous federated learning, where clients can update model parameters asynchronously based on their resource availability. This asynchronous nature allows more flexibility in the federated learning process, particularly in resource-constrained environments. In addition, DAG-DLT is specifically designed for scenarios such as the IoT and can be better adapted to edge computing environments. These features of DAG-DLT make it a suitable choice for decentralized FL in edge device environments where resource limitations and connectivity constraints are prevalent. Among the existing studies \cite{lu2020b,schmid2020,yuan2021,beilharz2021,cao2021}, an Implicit Model Specialization FL called Specializing DAG FL (SDAFGL) \cite{beilharz2021} proposed by Beilharz et al. is attention-gaining. SDAFGL focuses on improving model accuracy locally on the client side. It employs an accuracy-based biased random walk tip selection algorithm, which enables implicit specialization of the model. This approach exhibits good performance, particularly on clustered-wise non-iid datasets.

\section{DAG-ACFL}
\label{sec:framework}
\subsection{Motivation}
\begin{figure*}[h]
    \centering
\includegraphics[width=0.9\textwidth]{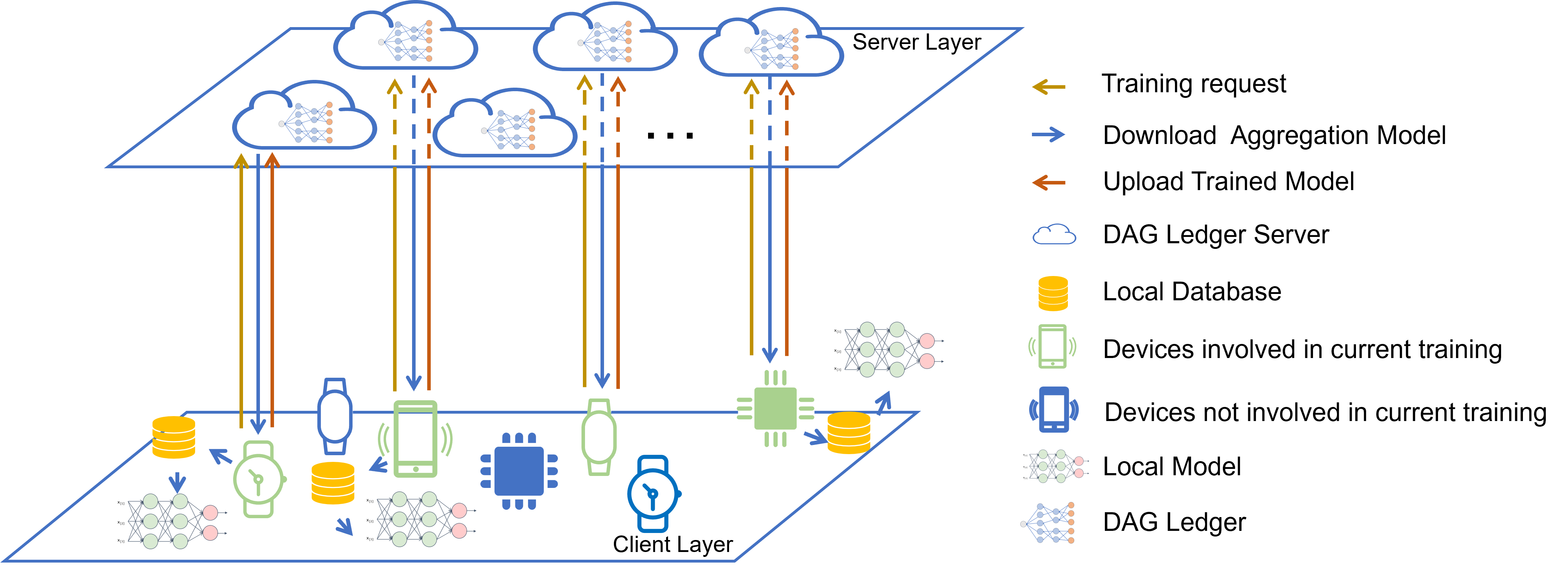}
    \caption{An overview of DAG-ACFL. The DAG-ACFL framework comprises multiple FL server nodes, each maintaining a DAG Ledger, and clients with local train and test sets. The FL server nodes form a DLT system. The green devices represent participating clients in the training process, while the blue devices indicate clients not currently involved in training.}
    \label{fig:framework}
\end{figure*}

To better understand our work, we illustrate the motivation behind our research before introducing DAG-ACFL. Clustered FL has demonstrated significant advantages in addressing the challenges posed by clustered-wise non-IID data distributions compared to Vanilla FL. However, existing clustered FL approaches still follow the synchronous FL framework, which is challenged by scalability and security. On the one hand, synchronous FL requires the server to receive all model parameters from all selected clients participating in the current training round before aggregating them. That can result in the slowest client blocking the training process, particularly in edge device scenarios with limited network bandwidth and computing resources. Fully orchestrating the FL process through a central server becomes challenging. On the other hand, synchronous FL is vulnerable to single points of failure and malicious attacks. If the server fails or is attacked, the entire FL system can be paralyzed, potentially leading to privacy leaks and other security issues.

These challenges necessitate exploring alternative approaches that address synchronous FL's scalability and security concerns. SDAFGL seems an effective solution, but it has some drawbacks when deploying it in real-world edge device scenarios. Firstly, the decentralization aspect of the DLT requires storing a whole distribution ledger containing complete model parameters throughout the entire FL training process on each client. However, edge devices often have limited storage resources, making storing the entire distribution ledger on each client impractical, particularly when the number of participating clients is enormous. Secondly, in SDAFGL, device communication occurs through peer-to-peer communication, which can lead to increased communication overhead as the number of clients grows, which is challenging for edge devices, especially when network connectivity is unstable, or bandwidth is limited. Lastly, transmitting model parameters between clients in SDAFGL increases the risk of data leakage and security breaches. As the model parameters are shared among clients, there is a higher risk of information leakage.

Therefore, finding ways to implement decentralized asynchronous federated learning for clustered-wise non-IID data while considering the limited resources of edge devices is a significant research problem. Considering the abovementioned limitations and drawing inspiration from clustered FL and SDAFGL, we propose a bi-layer asynchronous clustered FL framework based on DAG-DLT.

\subsection{DAG-ACFL Framework Overview}
The overview of DAG-ACFL is shown in Figure \ref{fig:framework}. The DAG-ACFL framework is a bi-layer architecture consisting of two layers: the server layer and the client layer. It is important to note that the depiction in Figure \ref{fig:framework} is a simplified illustration, and the actual deployment may require adjustments based on specific circumstances.

In the {\bf{Server Layer}}, multiple DAG Ledger server nodes are deployed. These server nodes communicate with each other either through network broadcasting or peer-to-peer communication to update the DAG ledger, which ensures the synchronization of new transactions and information across the entire distributed ledger maintained by the DAG Ledger servers.

\begin{figure}[h]
    \centering
    \includegraphics[width=0.33 \textwidth]{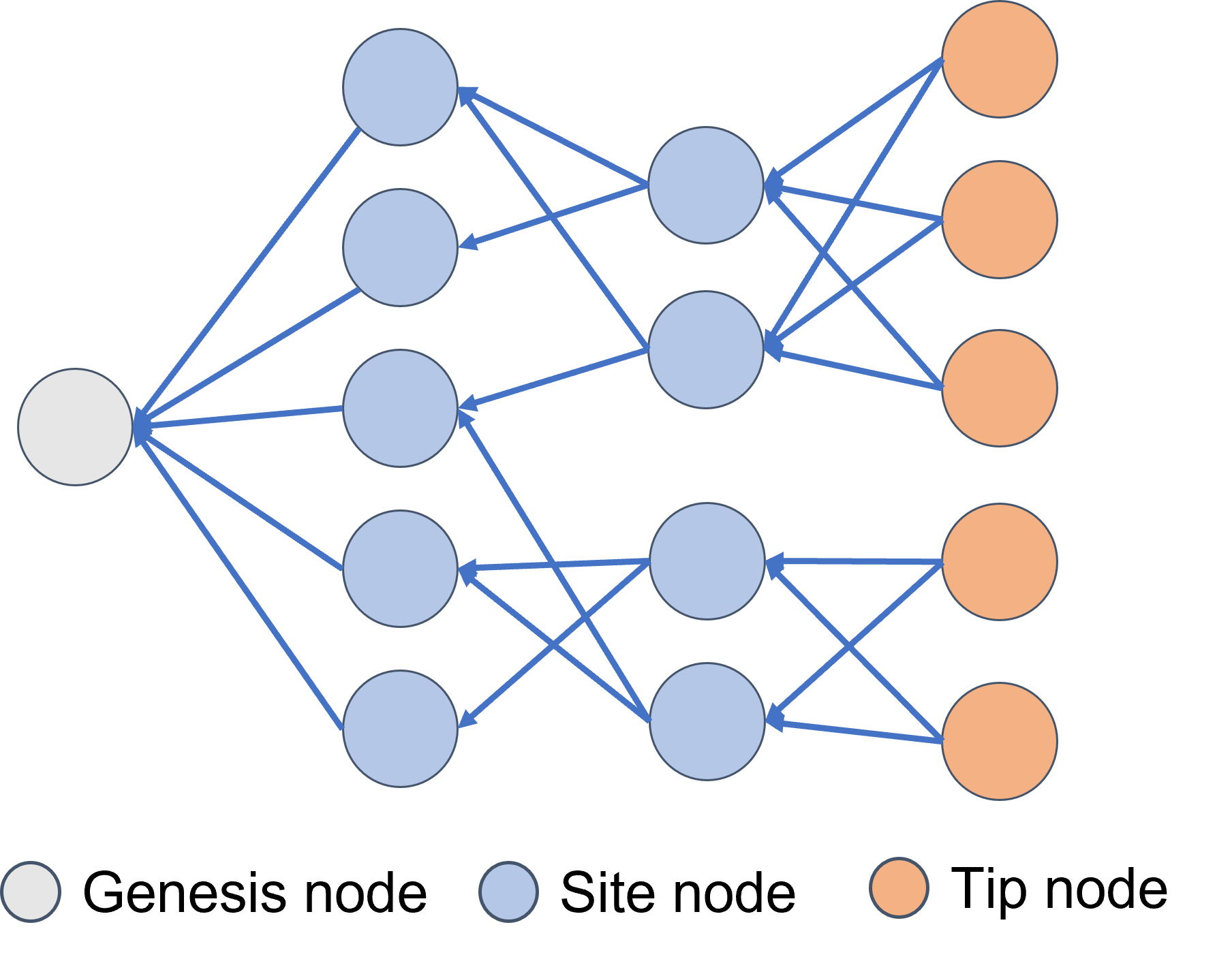}
    \caption{A diagram of the DAG distributed ledger. The nodes in the DAG Ledger can be divided into three categories: genesis nodes, site nodes, and tip nodes.}
    \label{fig:dag-dlt}
\end{figure}

Each DAG ledger server node is responsible for maintaining a synchronized DAG Ledger, which records the model parameters of each client participating in the training process. The structure of the DAG ledger can be visualized in Figure \ref{fig:dag-dlt}. The nodes within the DAG-DLT can be categorized into three types: (1){\bf{Genesis Node}}: The genesis node is the first node in the entire DAG ledger. It defines the initial state of the ledger, including parameters such as the type of tip selection algorithm, network settings, and other relevant information. (2){\bf{Site Node}}: Site nodes serve as the basic building blocks of the DAG ledger. Each site node is approved by at least two site nodes or tip nodes. Site nodes contain crucial information, including model parameters, timestamps, and hashes of parent nodes. (3){\bf{Tip Node}}: Tip nodes are particular site nodes that any other node within the DAG has not approved. They represent the latest additions to the ledger and await approval from other new transactions.

DAG ledger servers can be deployed in real-world edge device environments on multiple mobile edge computing cloud servers or communication base stations. This deployment strategy allows for closer proximity to the edge devices and facilitates efficient communication and model exchange between the clients and the DAG ledger servers.

The {\bf{client layer}} of the DAG-ACFL framework consists of multiple devices, including smartphones, sensor devices, and IoT devices. Each client within the client layer maintains its local model and dataset. The local model is initialized with random model parameters at the beginning of the training process. Due to the network conditions and the mobility of mobile edge devices, the connection between the clients and the DAG ledger servers is flexible. Clients can establish connections with different DAG ledger servers at any given time. They can select the closest and most suitable server based on network conditions and their location. This flexibility enables the DAG-ACFL framework to adapt to various mobile edge computing environments. By considering the dynamic network state and the mobility of clients, the DAG-ACFL framework ensures efficient and adaptable communication between clients and servers. This adaptability contributes to the overall effectiveness and performance of the federated learning process in different mobile edge computing scenarios.

Indeed, the DAG-ACFL framework offers flexibility in deployment across different environments, including the Internet of Vehicles (IoV) and Smart City scenarios. In the context of IoV, the DAG ledger server can be deployed on base stations such as Road Side Units (RSUs). The clients in this scenario would be various devices within vehicles participating in the federated learning process. These devices can include onboard computers, infotainment systems, or other connected vehicle components. As for Smart City environments, the DAG ledger server can be deployed on distributed nodes throughout the city. These nodes could be located on infrastructure elements such as lampposts, utility poles, or other central points within the city. The clients in this context would encompass various devices and nodes within the Smart City ecosystem, such as cameras, sensors, intelligent street lamps, environmental monitoring systems, and more. The versatility of the DAG-ACFL framework allows for its deployment and adaptation in diverse environments, catering to each scenario's specific requirements and characteristics. By leveraging the decentralized and asynchronous nature of DAG-DLT, the framework can facilitate efficient federated learning across different domains, enhancing collaboration and data privacy while accommodating the unique infrastructure and device compositions found in IoV and Smart City environments.

\subsection{Workflow of DAG-ACFL}
\label{sec:workflow}
\begin{figure}[h]
    \centering
    \includegraphics[width=0.5\textwidth]{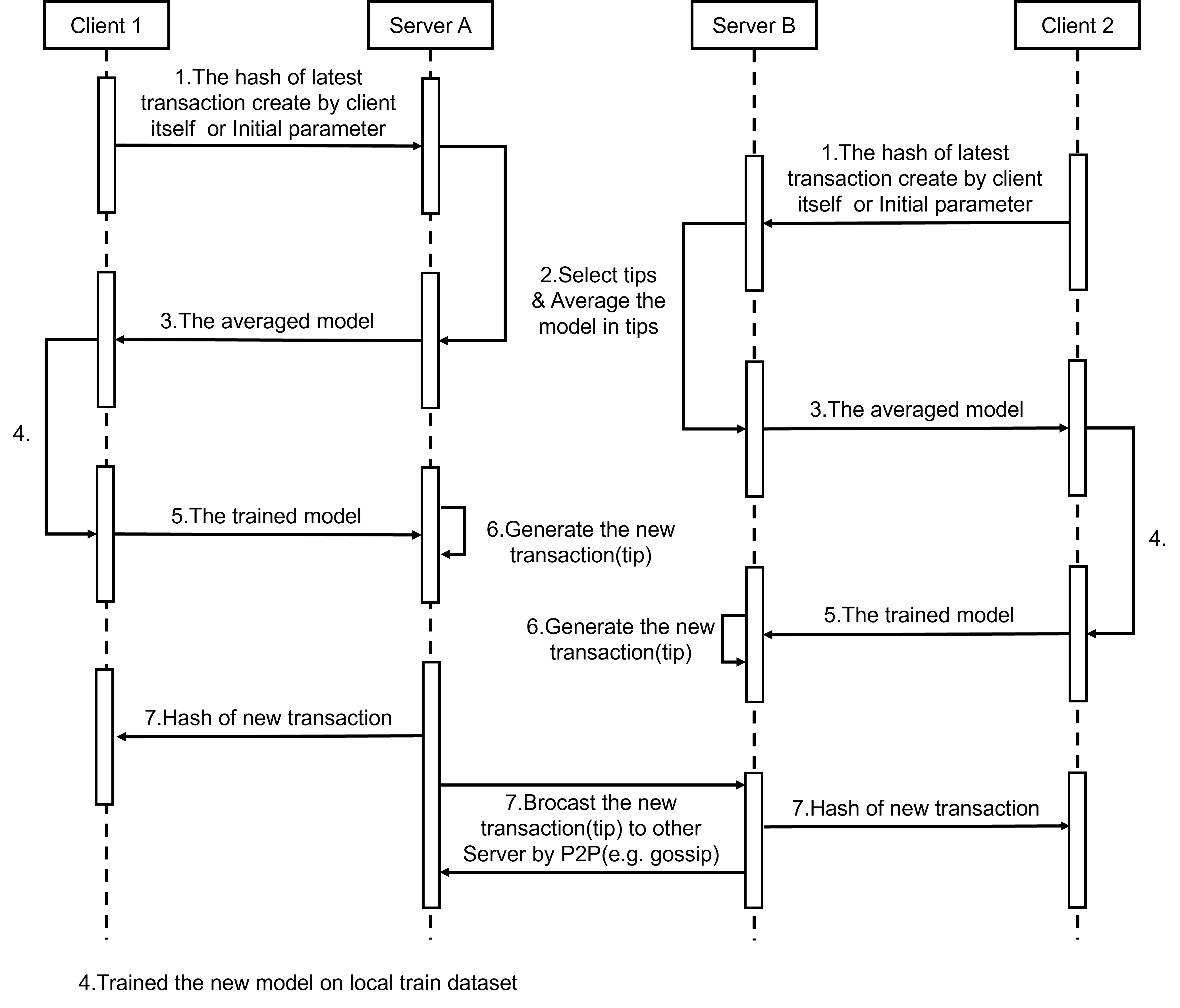}
    \caption{A timing diagram of DAG-ACFL. There are two clients, 1 and 2, and two DAG ledger servers, A and B. To show the asynchronous training of the DAG-ACFL. Each client trains the model using its local dataset. Client 1 is connected to the DAG ledger server A, and client 2 is connected to the DAG ledger server B. And the servers A and B are synchronized with each other after a certain period.}
    \label{fig:workflow}
\end{figure}

In this section, we will provide a detailed overview of the workflow of DAG-ACFL. DAG-ACFL is a client-oriented framework, and the workflow of DAG-ACFL differs from that of Vanilla FL. In DAG-ACFL, clients independently and periodically send requests to the server to obtain the latest model parameters. They then train their local models using their datasets without waiting for instructions from the central server. This asynchronous and decentralized training model enhances the flexibility and parallelism of the federated learning process. A timing diagram is used in Figure \ref{fig:workflow} to visualize the workflow of DAG-ACFL.

The training process of a client in DAG-ACFL can be divided into seven steps, as depicted in Figure \ref{fig:workflow}. Initially, client $i$ selects a DAG ledger server within its communication range and establishes a connection with the selected server. The client then transmits the hash value of the latest transaction it created to the server. The server verifies the validity of the hash value and queries the ledger using the hash value to obtain the corresponding model parameter from the ledger. The server then utilizes a predefined tip selection algorithm. It is worth noting that when a client joins the federated learning for the first time, it will obtain the model parameters stored in the genesis node and performs multiple epoch training using its local training dataset. Once the training is completed, the client returns the updated model parameter to the server. This process corresponds to step 1, as Figure \ref{fig:workflow} shows.

After the client sends hash values or its model parameter to the server, the server performs the tip selection algorithm (step 2 as illustrated in Figure \ref{fig:workflow}) to select the tip nodes. The model parameters stored in the selected tip nodes will be aggregated to obtain the new model $\omega_{agg}$ like FedAvg. The server then sends the $\omega_{agg}$ back to client $i$ (step 3 as illustrated in Figure \ref{fig:workflow}). We will detail the tip selection algorithm based on model parameters in Section \ref{sec:tsa}.

Upon receiving the updated model $\omega_{agg}$ from the server, client $i$ proceeds to train this model using its local dataset for multiple epochs. After training, the client returns the updated model parameter to the server (steps 4 and 5, as illustrated in Figure \ref{fig:workflow}).

The server will package the model parameter receive from the client $i$ into a transaction. The server then computes the hash value of the transaction and publishes it to the DAG ledger. This process is step 6, shown in Figure \ref{fig:workflow}. And then, the server will send the hash valid to the client $i$ for its next training and broadcast the new transaction and the approval information with the tip nodes to the other servers. This process is step 7, shown in Figure \ref{fig:workflow}.
 
In this manner, client $i$ completes a round of training within the DAG-ACFL framework. The overall process is outlined in Algorithm \ref{alg:dagacfl}.

\begin{algorithm}[htbp]
    \caption{DAG-ACFL}
    \label{alg:dagacfl}
    \KwIn{$\boldsymbol{\mathit{N}}$ clients with local datasets, $\boldsymbol{\mathit{D_i}}$ is the local train dataset hold in clients $\boldsymbol{\mathit{i}}$,$\boldsymbol{\mathit{E}}$ is number of local epochs, $\boldsymbol{\mathit{E'}}$ is number of local initial train, $\boldsymbol{\mathit{\eta}}$ is learning rate, hash of latest transaction created by $client_i$ is $\boldsymbol{\mathit{hash_i}}$, DAG Ledger $\boldsymbol{DAGL}$, number of tips $\boldsymbol{nt}$}
    \KwOut{Optimized local model $\omega$ }
     \For{each client $i \in |N|$ {\bf{in parallel}}}{
        $hash_i \leftarrow 0$ \tcp{Initialize $\boldsymbol{\mathit{hash_i}}$}
        
        Client $i$ establish the connection with DAG Ledger server\;
        \If{$hash_i == \phi $}
        {   
            Random initial local model parameter $\omega_i$ of clients $i$\;
            $\omega_i \leftarrow$ {\bf{ClientLocalTrain}}($\omega_i, E', \eta$)\;
            Transmit $\omega_i$ to the currently connected DAG distributed ledger {\bf{server}}\;
        }
        \Else
        {   
            Transmit the $\boldsymbol{\mathit{hash_i}}$ to the currently connected DAG ledger {\bf{server}}\;
            The {\bf{server} }obtains $\omega_i$ from the DAG ledger according $\boldsymbol{\mathit{hash_i}}$\;
        }
        $\omega_{agg} \leftarrow$ {\bf{ServerTipAggregate}}($DAGL$, $S_i$, $nt$)\;
        $\omega_i \leftarrow$ {\bf{ClientLocalTrain}}($\omega_{agg}, E, \eta$)\;
        Transmit $\omega_i$  to the currently connected DAG distributed ledger {\bf{server}}\;
        Receive the $\boldsymbol{\mathit{hash_i}}$ of the transaction from the currently connected {\bf{server}}.\;
    }

    \SetKwFunction{ClientLocalTrain}{ClientLocalTrain}
    \SetKwProg{Function}{Function}{:}{end}
    \Function{{\bf{ClientLocalTrain}}{($\omega, E, \eta$)}}{
        \For{$e=1$ \KwTo $E$}{
            $\omega \leftarrow \omega - \eta \nabla L(b;\omega)$ for local batch $b \in B$\;
        }
        \Return{$\omega$}\;
    }

    \SetKwFunction{ServerTipAggregate}{ServerTipAggregate}
    \SetKwProg{Function}{Function}{:}{end}
    \Function{\bf{ServerTipAggregate}{($DAGL$, $\omega_i$, $nt$)}}{
        tips = {\bf{TSA}}($\omega_i$,$num\_tips$)\;
        $\omega \leftarrow$ Aggregate the model in each tip with FedAvg\;
        \Return{$\omega$}\;
    }
\end{algorithm}

\subsection{Tip selection algorithm based on similarity of model parameters}
\label{sec:tsa}
The tip selection algorithm is the cornerstone of DAG-ACFL. In this section, we introduce a tip selection algorithm based on the similarity of model parameters. An effective tip selection algorithm can accurately identify tip nodes created by clients with similar data distributions enabling the clustering of transactions generated by clients with similar dataset distributions, leading to better adaptation of the model to the varying data distributions across clients. Therefore, selecting tip nodes that suit the client's data distribution is crucial for ensuring the training effectiveness of DAG-ACFL.

Model parameters are the fundamental components of FL and play a crucial role in the FL process implementation. As the data distribution of each client within each superclass differs, the model parameters trained using different datasets also vary. And the parameters in the later layers of the model can capture differences in data distribution better than the parameters in the earlier layers. Thus, we can use the similarity of model parameters in the latest layer to measure the similarity of data distribution among clients. Based on this, we propose a tip selection algorithm based on the similarity of model parameters.  

The core idea behind this algorithm is to compute the cosine similarity between the model parameters of the tip nodes and those of the client $i$. The similarity between the model parameters of client $i$ and the model parameters of the tip node can be defined as follows.
\begin{equation}
    \label{eq:similarity}
    \begin{aligned}
        \text{similarity} &= \cos(\omega_i, \omega_{\text{tip}}) \\
        &= \frac{<\omega_i , \omega_{\text{tip}}>}{\|\omega_i\| \|\omega_{\text{tip}}\|}
    \end{aligned}        
\end{equation}

By calculating the cosine similarity between the model parameters, we obtain a value in the range of [-1, 1] that indicates the similarity between $\omega_i$ and $\omega_{tip}$. A value closer to 1 indicates a higher similarity between $\omega_i$ and $\omega_{tip}$, while a value closer to -1 indicates a lower similarity between $\omega_i$ and $\omega_{tip}$.

Next, we select the tip nodes with the highest similarity to aggregate the model parameters within these tip nodes, resulting in $\omega_{agg}$. The algorithm for this process is presented in Algorithm \ref{alg:tsa}. When receiving a request from client $i$, the server first obtains all the tip nodes in the DAG ledger (Line 1 of Algorithm \ref{alg:tsa}). It then calculates the similarity of model parameters between each tip node and the newest model of client $i$ (Lines 2-5 of Algorithm \ref{alg:tsa}). Finally, the models from the top $nt$ tip nodes with the highest similarity are selected, and the aggregated models are returned to the client using the FedAvg algorithm (Lines 6 and 7 of Algorithm \ref{alg:tsa}).

\begin{algorithm}[h]
    \caption{Tip selection algorithm based on similarity of model parameters}
    \label{alg:tsa}
    \KwIn{DAG Ledger $\boldsymbol{DAGL}$, $\omega_i$ is the model parameters of client $i$, $\omega_{tip}$ is the model parameters in a tip node, number of tips $nt$}
    \KwOut{Selected tips}
    \tcp{Get all tips in DAG Ledger}\
    tips $\leftarrow ${\bf{GetAllTips}}($DAGL$)\;
    \tcp{Initialize similarities list}\
    $S \leftarrow$ $\emptyset$\; 
    \For{tip in tips}
    { 
        S[tip] $\leftarrow$ Computed by Eq.(\ref{eq:similarity})\;
    }
    $stips \leftarrow$ {\bf{TopK}}(S, $nt$)\;
    \Return{$stips$}\;
\end{algorithm}

\subsection{Adaptive tip selection algorithm}
In the tip selection algorithm proposed in Section \ref{sec:tsa}, the number of tips selected by the server is predefined and fixed.  However, the number of tips that satisfy the condition varies for different clients; and varies from client to client. Therefore, we propose an adaptive tip selection algorithm to select the number of tips for each client dynamically.

\begin{figure}[h]
    \centering
    \includegraphics[width=0.4\textwidth]{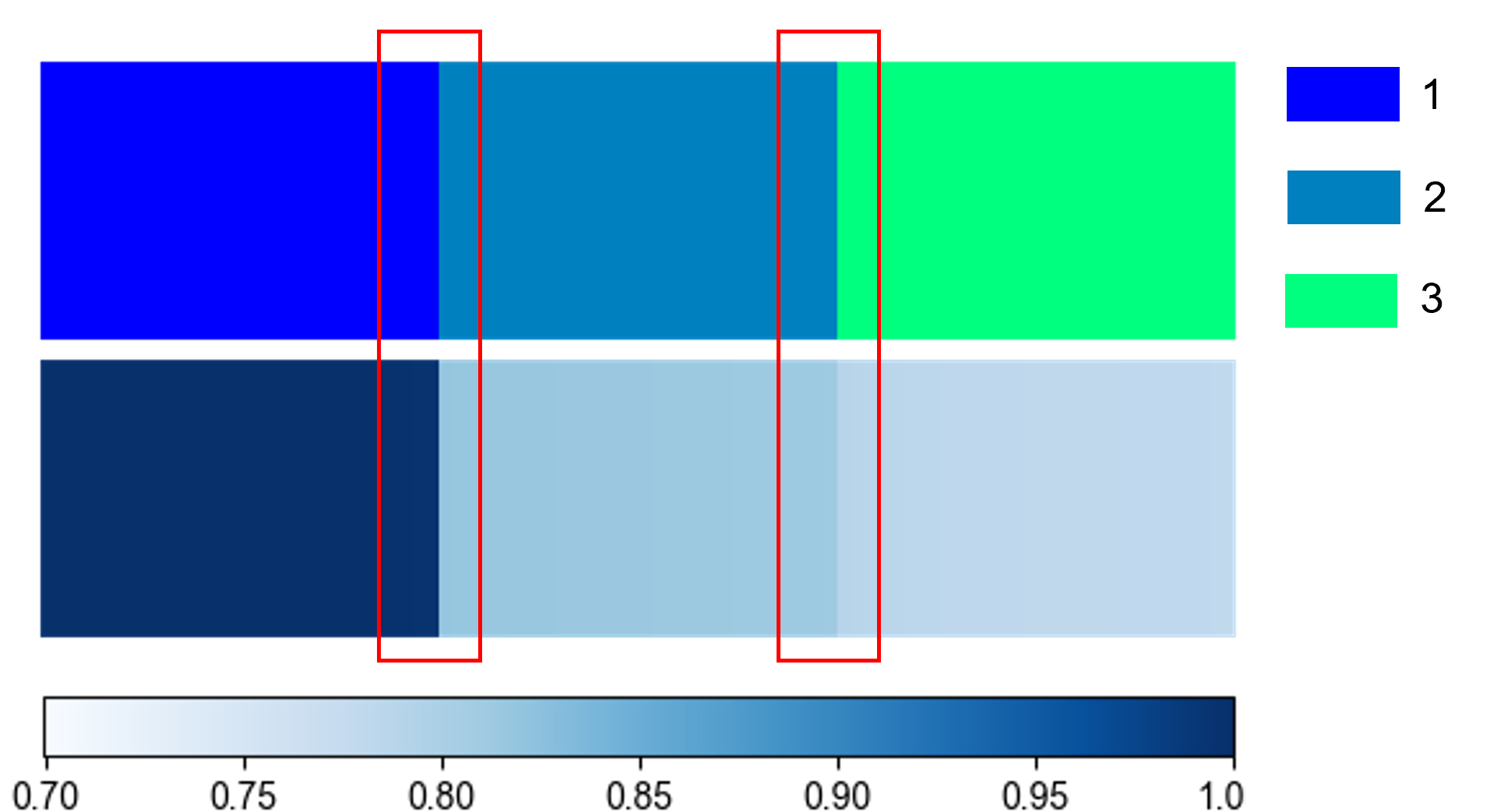}
    \caption{We randomly selected the sequence of model parameter similarities computed between the model parameters of client $i$ and the model parameters of the tip nodes published by other clients. We then ranked the tip nodes based on their similarity values. The upper represents the cluster IDs of the clients who published the tip nodes, while the lower heatmap represents the sorted sequence of similarities.} 
    \label{fig:heatmap}
\end{figure}

We conducted a simple experiment to illustrate better the core idea of our adaptive tip selection algorithm. We divided the MNIST dataset into three superclasses, each containing 30 clients. We trained a Multi-Layer Perceptron (MLP) model with one hidden layer of 128 neurons. Next, we randomly selected a client $i$ and obtained the similarity sequence in a randomly chosen training round. We then visualized the similarity sequence as a heatmap, shown in Figure \ref{fig:heatmap}.

The heatmap clearly illustrates substantial differences in similarity values between nodes from different clusters. Furthermore, there is a noticeable change point in the similarity sequence. Therefore, we only need to find the first change point to find the tip nodes with similar data distribution to the client.

\begin{figure}[h]
    \centering
    \includegraphics[width=0.45\textwidth]{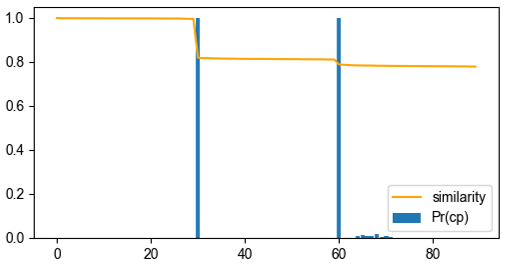}
    \caption{The result of change point detection for the similarity sequence. The orange line represents the similarity sequence, while the blue bars represent the probabilities (confidence level) computed by BEAST for each point to be a change point. The horizontal axis represents the index of the similarity sequence, while the vertical axis represents the value of the similarity sequence and the probability of a change point.} 
    \label{fig:beast}
\end{figure}

To adaptively identify the most appropriate change point, we introduce the Bayesian Estimator of Abrupt change, Seasonality, and Trend (BEAST) algorithm \cite{zhao2019}, a change point detection algorithm, into our DAG-ACFL. BEAST can divide time series into multiple segments based on detected change points. It has demonstrated excellent performance in various areas, including change point detection, trend analysis, time series decomposition, and time series segmentation. We apply the BEAST algorithm to detect the change point in the similarity sequence depicted in Figure \ref{fig:heatmap}. The result of the change point detection using BEAST is visualized in Figure \ref{fig:beast}.

Figure \ref{fig:beast} demonstrates that BEAST has successfully detected the change point and the change point with a high confidence level. The first change point is what we need. Based on this result, we can utilize BEAST to detect the change point in the similarity sequence; and select the tip nodes in front of the first change point with a high confidence level. The algorithm for this process is presented in Algorithm \ref{alg:atsa}.
\begin{algorithm}[h]
    \caption{Adaptive tip selection algorithm based on similarity({\bf{ATSAS}})}
    \label{alg:atsa}
    \KwIn{DAG Ledger $\boldsymbol{DAGL}$, $\omega_i$ is the model parameters of client $i$, $\omega_{tip}$ is the model parameters in a tip node, the threshold of the confidence level is $\alpha$, number of minimum tips $nt$}
    \KwOut{Selected tips} 
    tips $\leftarrow ${\bf{GetAllTips}}($\boldsymbol{DAGL}$)\;
    $S \leftarrow$ $\emptyset$\; 
    \For{tip in tips}
    { 
        $Similarity[tip] \leftarrow$ Computed by Eq.(\ref{eq:similarity})\;
    }
    \tcp{Sort similarities and tips according to similarities in descending order}\
    sort($Similarity$,tips)\;
    \tcp{compute the confidence level of point in $Similarity$ }\
    $Confidence \leftarrow$ {\bf{BEAST}}($Similarity$)\;
    \For{$i in  range(len(Confidence))$}
        {\If{$Confidence[i] > \alpha$}
            index $\leftarrow$ $i-1$\;
            {break\;}
        }
    \If {$index < nt$}
        {
            index $\leftarrow$ $nt$\;
        }
    \Return{tips[0:index]}\;
\end{algorithm}

In Algorithm \ref{alg:atsa}, the first step is to calculate the model parameters' similarity between the client and the tip nodes in the DAG ledger, resulting in a sequence of similarity values $Similarity$. Subsequently, $Similarity$ is sorted in descending order. Then, the BEAST algorithm is applied to detect the change point in $Similarity$, yielding the confidence level sequence $Confidence$ of each point. We determine the index $index$ of the first change point in $Confidece$ where the confidence level exceeds the threshold $\alpha$ (Lines 8-14 in Algorithm \ref{alg:atsa}). 

Finally, we select the former $index$ tip nodes. However, at the early stages of model training, the change point in similarity may be insignificant. To address this issue, we set a minimum number of tip nodes, denoted as $nt$, to be selected. If the number of tip nodes selected is less than $nt$, we will choose the first $nt$ tip nodes instead (Lines 15-17 in Algorithm \ref{alg:atsa}).

In summary, DAG-ACFL implements asynchronous FL and implicit clustering of similar clients using a tip selection algorithm based on the similarity of model parameters combined with a DAG ledger. Multiple DAG ledger servers are deployed within the DAG-ACFL framework to achieve the distribution and scalability of FL. During each communication, clients only need to send the hash value of the latest transaction to the server, except for their initial participation in FL. This approach minimizes additional communication overload. The key features of our DAG-ACFL framework are as follows: 
\begin{itemize} 
    \item The DAG-ACFL is a bi-layer FL framework based on DAG-DLT, it can reduce the communication and the storage consumption of the client compare with the SDAGFL. The DAG-ACFL framework can be deployed in various edge device scenarios, including IoV and Smart City environments.
    \item The DAG-ACFL is a client-oriented framework. The client can join the FL at any time and leave the FL at any time. It is an asynchronous and decentralization framework that fully uses the advantage of DAG-DLT.
    \item The clients are implicitly clustered according to the similarity of the model parameters based tip selection algorithm. The clients with similar data distribution are clustered into the same cluster without predefined the number of clusters.
\end{itemize} 

\section{Evaluation}
\label{sec:evaluation}
In this section, we will present the experimental results of DAG-ACFL using a PyTorch-based simulation implementation. We constructed cluster-wise datasets using the MNIST \cite{deng2012} and CIFAR-10 \cite{krizhevsky2009} datasets. We will showcase the performance of the DAG-ACFL framework on these datasets and demonstrate the effectiveness of implicit client clustering and the adaptive tip selection algorithm.

\subsection{Experimental Setup}
We conducted evaluations of our DAG-ACFL framework on two datasets: MNIST and CIFAR-10, each with its corresponding models. The client datasets were split into training and testing sets using an 80:20 ratio. And for the CNN model, we use the last two layers of the model for similarity measures. To assess the performance of our framework, we compared it with five baselines: FedAvg \cite{mcmahan2017}, FlexCFL \cite{duan2021}, IFCA \cite{ghosh2022a}, FeSEM \cite{long2023}, and SDAFGL \cite{beilharz2021}. The details are as follows.

{\it{Dtasets and Models}}

\begin{figure}[h]
    \centering
    \includegraphics[width=0.5\textwidth]{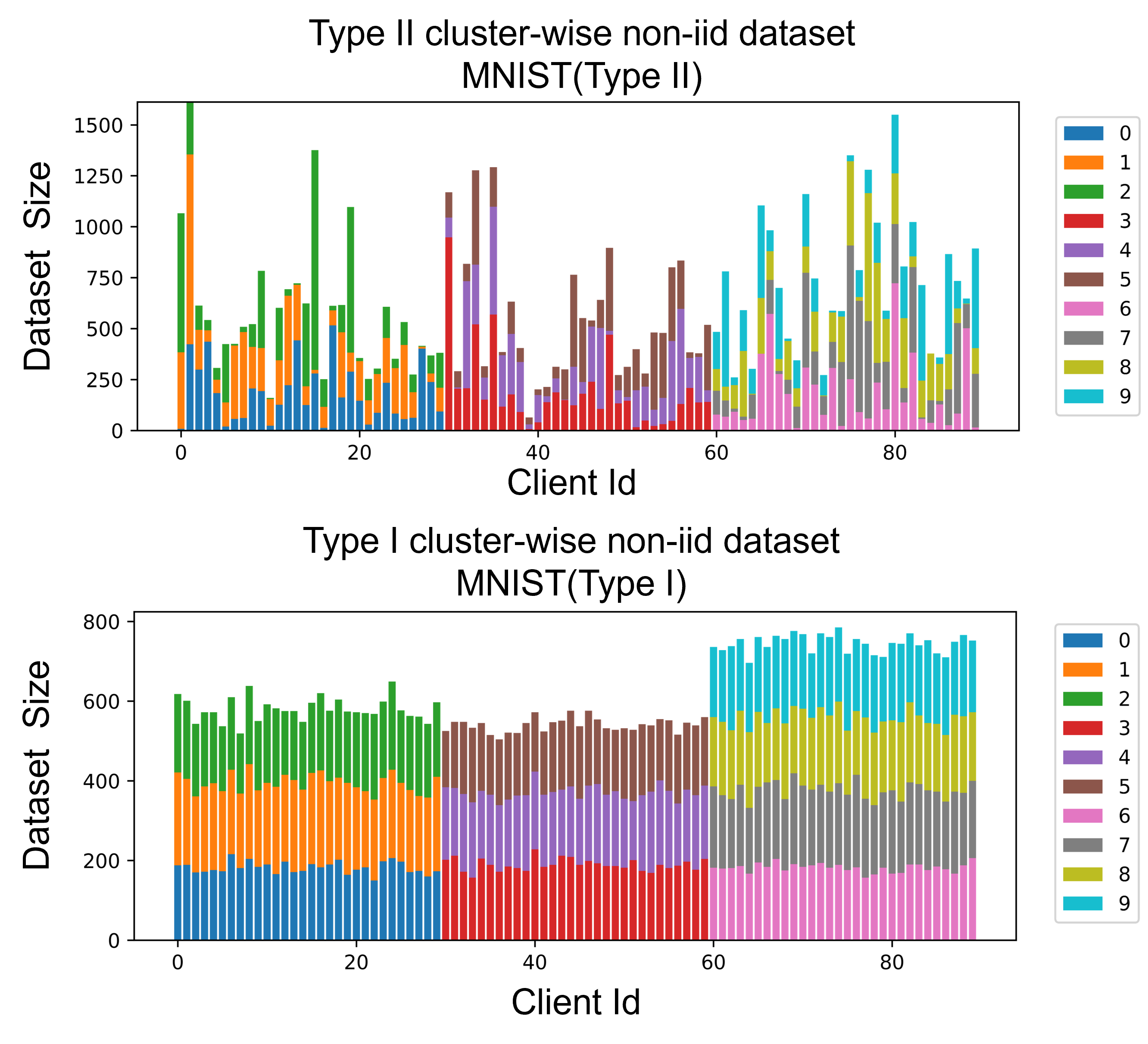}
    \caption{Distribution of client label with two settings of MNIST dataset. The different label is shown with a different color. It can be seen that the type of the client's label in each cluster is the same either in the type I setting or in the type II setting. But the number of each label in the type II setting is different.}
    \label{fig:mnist-distribute}
\end{figure}

For each dataset, take MNIST as an example, and we generated two types of data distributions. These distributions were based on labels among different clients within each superclass. In the type I dataset, the labels among the client in the same cluster are iid. In the type II dataset, the labels among the client in the same cluster are non-iid. Figure \ref{fig:mnist-distribute} presents the above two type label distribution of the MNIST dataset.

\begin{itemize}
    \item {\bf{MNIST}}: The MNIST dataset consists of 28$\times$28 pixel grayscale images of handwritten digits ranging from 0 to 9. We divided the dataset into three disjoint superclasses: \{0, 1, 2\}, \{3, 4, 5\}, and \{6, 7, 8, 9\}, with each superclass containing 30 clients. We generated two datasets, MNIST(type I) and MNIST(type II). For training, we used three models: logistic regression (LR), a multi-layer perceptron (MLP) with a hidden layer of 128 neurons, and a convolutional neural network (CNN). The CNN model consists of two convolutional layers with ReLU activation functions. Each convolutional layer has a kernel size of 5 and uses 32 filters. A maximum pooling layer with a pool size and stride length 2 follows both convolutional layers. In addition, the model also includes two fully connected layers with 2048 and 10 neurons, respectively. 
    \item {\bf{CIFAR-10}}: The CIFAR-10 dataset comprises 32$\times$32 pixel RGB images belonging to ten categories, including airplane, automobile, bird, cat, deer, dog, frog, horse, ship, and truck. In our experiment, we divided the dataset into four disjoint superclasses: \{airplane, automobile\}, \{bird, cat, deer\}, \{dog, frog, horse\}, and \{ship, truck\}. Each superclass consists of a different number of clients, specifically \{20, 30, 30, 20\} clients. Based on these superclasses, we generated two datasets, CIFAR-10(iid) and CIFAR-10(non-iid). For training, we used a CNN model. The CNN model consists of two convolutional layers with ReLU activation functions. Each convolutional layer has a kernel size of 5, and the first layer uses 32 filters, while the second layer uses 64 filters. A maximum pooling layer with a pool size and stride length 2 follows each convolutional layer. Finally, there are two fully connected layers with 512 and 10 neurons, respectively.
\end{itemize}

{\it{Baselines}}

\begin{itemize}
    \item FedAvg\cite{mcmahan2017}: This is a vanilla Federated Learning (FL) framework where clients perform local training and send their model updates to a central server.
    \item FlexCFL\cite{duan2021}: FlexCFL is a clustered FL framework that assigns clients to clusters based on a predefined number of clusters. It uses the Euclidean distance of Decomposed Cosine similarity (EDC) to measure similarity. In our simulation experiments, we set "RAC" and "RCC" to False and do not use inter-cluster clustering. Additionally, we set the pre-training epochs of each selected client in the group cold start and client cold start stage as shown in Table \ref{tab:hyperparameters}
    \item IFCA\cite{ghosh2022a}: IFCA is a clustered FL framework that assigns clients to the most appropriate cluster by iteratively testing the loss of the model across all clusters for each client.
    \item FeSEM\cite{long2023}: FeSEM is a framework that clusters clients by initializing multiple models in the server and minimizing the $l_2$ distance between the client model and the cluster model.
    \item SDAGFL\cite{beilharz2021}:  SDAGFL is a DAG-DLT based FL framework. It utilizes an accuracy-based biased random walk tip selection algorithm to find models that perform well on local data, enabling implicit model specialization and client cluster. In our simulation experiment, we aim to evaluate the performance of the accuracy-biased random walk tip selection algorithm, so we publish all new transactions to the DAG ledger instead of only publishing the transactions that perform better on the ``reference model'' selected by the client after each training. 
\end{itemize}

{\it{Metrics}}

DAG-ACFL is a client-oriented framework. To ensure a fair comparison with the baselines, we introduce the concepts of rounds and a global model similar to Vanilla FL. In DAG-ACFL, each client participates in training at most once during a specific period referred to as a round. The global model for client $i$ is the aggregate model obtained from all the selected tip nodes in the DAG ledger. 

Our experiment provides each client in these frameworks with train and test datasets. We utilize the test dataset to evaluate the performance of the global model. In the case of FedAvg, the global model is evaluated using the local datasets of all clients. For FlexCFL, IFCA, and FeSEM, the group model is evaluated using the test datasets of the clients within each respective group. In SDAGFL, the test clients are all the clients historically partitioning in the FL, and each client its tip slection algorithm and average the model of the selected tips as the global model and evaluates it using its local test dataset.  In the case of DAG-ACFL, each client that has participated in the training selects its global model and evaluates it using its local test dataset. 

To measure the overall performance, we compute the average of the evaluation results across all clients. In FlexCFL, IFCA, FeSEM, SDAGFL, and DAG-ACFL, the test clients used for evaluating the group model include all clients historically assigned to the respective group. For FedAvg, the performance of all participating clients is tested. As for SDAFGL, we use its own evaluation method. By averaging the evaluation results across clients, we obtain an overall measure of the framework's performance. We fix the random seeds of the clients' selection for all frameworks to ensure a fair comparison and reproducibility.
 
Table \ref{tab:hyperparameters} shows the training hyperparameters settings. The optimizer stochastic gradient descent (SGD) is used.
\begin{table}[h]
    \centering
    \caption{Hyperparameters Settings} 
    \begin{tabular}{lcc}
      \toprule
      {\bf{Hyperparameters}} & MNIST & CIFAR-10 \\
      \midrule
      {\bf{Training rounds}} & 200 & 200 \\
      {\bf{Participation rate $\gamma$} }& 1(0.5) & 1(0.5) \\
      {\bf{Local epochs $E$}} & 1 & 5 \\
      {\bf{Batch size $b$}} & 10 & 10 \\
      {\bf{Learning rate $lr$}} & 0.05 & 0.001 \\
      {\bf{Pre-training epochs}} & 1 & 20 \\
      \bottomrule
    \end{tabular}
    \label{tab:hyperparameters}
\end{table}

\subsection{Comaparisons with Baselines}
In this section, we present the results of our evaluation, comparing DAG-ACFL with the baselines to highlight the advantages of our framework. We assess the performance of DAG-ACFL and the baselines using different models and data distributions on the MNIST and CIFAR-10 datasets. We consider two client participation rates: 1 and 0.5, indicating that all clients and half of the clients participate in the training process, respectively. Since clustered FL methods such as IFCA, FeSEM, and FlexCFL require a predefined number of clusters, we set the number of clusters to 3 for the MNIST dataset and 4 for the CIFAR-10 dataset. Additionally, we conduct additional experiments with the number of clusters set to 2 for both the MNIST and CIFAR-10 datasets and evaluate with FlexCFL. By evaluating the performance of DAG-ACFL and baselines under various scenarios, we aim to demonstrate the effectiveness and advantages of our framework in handling different models and data distributions.

\begin{figure*}
	\centering
	\includegraphics[width=\textwidth]{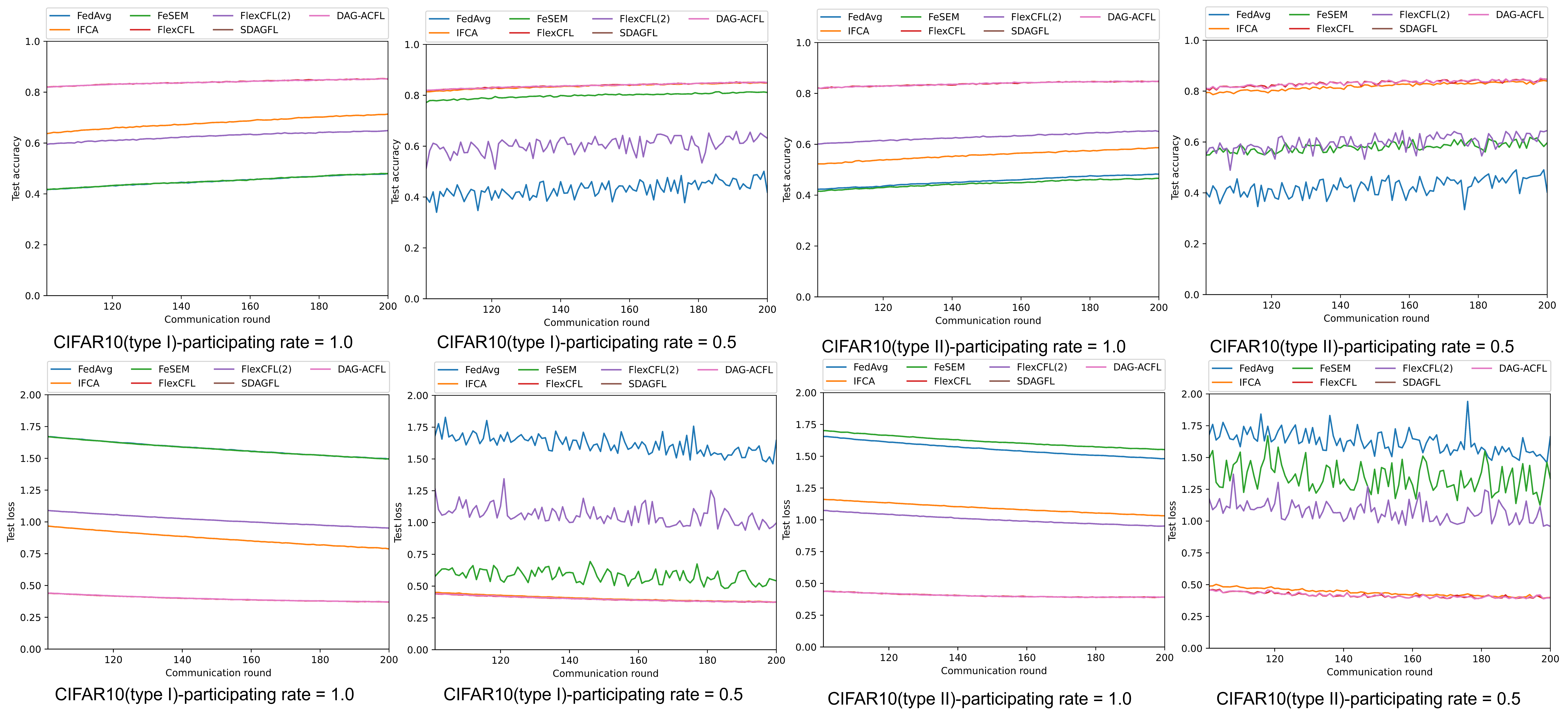}
	\caption{The test accuracy and loss from round 100 to 199 of the above frameworks on the CIFAR-10 dataset. The upper figures are the test accuracy, and the lower is the test loss. The figures from left to right are the results of CIFAR10(type I) and CIFAR10(type II) with participation rates of 1.0 and 0.5, respectively.}
	\label{fig:performance}
\end{figure*}

The experimental results are shown in Table \ref{tab:performance} and Figure \ref{fig:performance}. Figure \ref{fig:performance} shows the test accuracy and loss on CIFAR10(type I) and CIFAR10(type II)  datasets with different clients' participation rate. 
\begin{table*}[h]
    \centering
    \caption{Comaparisons with Fedavg\cite{mcmahan2017}, FlexCFL\cite{duan2021}, IFCA\cite{ghosh2022a}, FeSEM\cite{long2023} and SDAGFL\cite{beilharz2021} on MNIST and CIFAR-10 with two different label distribute setting in each superclass.} 
    \begin{tabular}{lccccccc}
      \toprule
      {\bf{Dataset-Model}}    &\makecell{Participation\\ rate}   & Fedavg & IFCA & FeSEM & FlexCFL & FlexCFL(2) &   \makecell{DAG-ACFL} \\
      \midrule
        {\bf{MNIST(type I)-LR}} &        &0.8791 &0.8862 &0.8862 &0.9726 &0.9426    &0.9726  \\
        {\bf{MNIST(type II)-LR}}&     &0.8827 &0.8890 &0.8887 &0.9721 &0.9411   &0.9722   \\
        {\bf{MNIST(type I)-MLP}} &       &0.9718 &0.8588 &0.8588 &0.9726 &0.9385    &0.9728   \\
        {\bf{MNIST(type II)-MLP}} &   &0.8310 &0.8393 &0.8356 &0.9721 &0.9398   &0.9716   \\
        {\bf{MNIST(type I)-CNN}}& 1.0    &0.9894 &0.9897 &0.9897 &0.9970 &0.9956     &0.9970   \\
        {\bf{MNIST(type II)-CNN}}&    &0.9897 &0.9899 &0.9895 &0.9975 &0.9957     &0.9976   \\
        {\bf{CIFAR10(type I)-CNN}}&      &0.4797 &0.7138 &0.4803 &0.8533 &0.6489     &0.8533   \\
        {\bf{CIFAR10(type II)-CNN}}&  &0.4830 &0.5865 &0.4664 &0.8475 &0.6531     &0.8488   \\
     \midrule 
        {\bf{MNIST(type I)-LR}} &        &0.8868 &0.9724 &0.9289 &0.9727 &0.9414    &0.9723    \\
        {\bf{MNIST(type II)-LR}}&     &0.8890 &0.9444 &0.9118 &0.9720 &0.9410    &0.9716  \\
        {\bf{MNIST(type I)-MLP}} &       &0.8747 &0.9692 &0.9473 &0.9722 &0.9409    &0.9727    \\
        {\bf{MNIST(type II)-MLP}} &   &0.8753 &0.9705 &0.9099 &0.9718 &0.9369     &0.9713    \\
        {\bf{MNIST(type I)-CNN}}& 0.5    &0.9897 &0.9891 &0.9915 &0.9973 &0.9959     &0.9973    \\
        {\bf{MNIST(type II)-CNN}}&    &0.9908 &0.9940 &0.9918 &0.9976 &0.9957     &0.9973    \\
        {\bf{CIFAR10(type I)-CNN}}&      &0.5011 &0.8504 &0.8132 &0.8524 &0.6580     &0.8521    \\ 
        {\bf{CIFAR10(type II)-CNN}}&  &0.4904 &0.8419 &0.6184 &0.8476 &0.6448     &0.8507   \\ 
      \bottomrule
    \end{tabular}
    \label{tab:performance}
\end{table*}

The experimental results show that all the other frameworks obtain performance improvements on cluster-wise datasets compared to the FedAvg FL framework, indicating that these frameworks are better for cluster-wise datasets. Compared with the IFCA, FlexCFL, SDAFGL, and DAG-ACFL methods, the FeSEM method has the worst performance caused by FeSEM's inability to cluster clients well \cite{duan2021}. Our proposed DAG-ACFL using an adaptive tip selection algorithm achieves comparable performance with FlexCFL. Although our experiment results are a little worse than FlexCFL in some experiment settings, it is negligible due to the small gap.

Notably, experimental results on the CIFAR10 dataset show that our method performs better on the cluster-wise dataset when the label distribution of clients within the same category is non-iid, or the client participation rate is 0.5. It is because our method focuses on selecting nodes with higher similarity. By analyzing the training process, we find that in the initial stage of model training, the clients do not choose entirely all the nodes released by the clients in the same class but select a portion of nodes with higher similarity, whose label distribution is more similar to that of the local clients, and thus the initial training direction of the model is more in line with the regional data distribution, thus improving the subsequent performance. Therefore, the initial training direction of the model is more in line with the local data distribution, which enhances the following performance.

Our experimental results also demonstrate an interesting phenomenon, i.e., FeSEM and ICFA outperform when the client participation rate is 0.5 than when the participation rate is 1.0. It is because the clustering effect of FeSEM and ICFA on clients does not precisely match the actual situation. A high participation rate interferes with the model. At the same time, moderate sampling can be regarded as the introduction of random discard in model training, which can suppress the noise of the client training set and improve the robustness of the model. Too many clients participate in the model instead of disturbing the model, and reducing the participation rate of clients is helpful for fusing the information.

The above experimental results show that our DAG-ACFL framework using the adaptive tip selection algorithm can adapt to the cluster-wise datasets and achieve comparable performance with the state-of-the-art methods. What's better is that our DAG-ACFL framework needs not to know the cluster information of clients in advance, which is more practical in real-world applications.

\subsection{Clustering evaluation of DAG-ACFL}
Since the DAG ledger in DAG-ACFL primarily displays the approval of different transactions rather than explicitly indicating client clustering, we need to assess the clustering effect using alternative approaches. 

Since we can access the client IDs in our experiments, we can transform the transaction approvals recorded in the DAG ledger into an adjacency matrix, denoted as $G$, representing client connectivity relationships. The edge weights in the adjacency matrix are determined by the total number of transactions generated by client $i$ that approve the transactions generated by client $j$. By constructing this adjacency matrix, we can quantify the connections and interactions between clients in DAG-ACFL. It enables us to analyze clients' relationships and clustering tendencies based on their transaction approvals. Thus, we introduce the modularity\cite{newman2004}, which is the same as the SDAGFL, to evaluate the clustering effect of DAG-ACFL. This evaluation helps us gauge how well DAG-ACFL accomplishes implicit client clustering.

{\it{Modularity}} is an important measure that helps evaluate the effectiveness of community detection algorithms and the overall structure of a network in the community discovery domain. Higher modularity values indicate a more robust community structure, where nodes within communities are densely connected, and connections between communities are relatively sparse. Equation \ref{eq:modularity} represents the formula used to compute modularity:
\begin{equation}
    Q=\frac{1}{2m}\sum_{i,j}[A_{ij}-\frac{k_ik_j}{2m}]\delta(c_i,c_j)
    \label{eq:modularity}
\end{equation}

In this equation, $A_{ij}$ represents the connection between nodes $i$ and $j$, while $k_i$ represents the degree of node $i$. The variable $m$ denotes the total number of edges in the network. The indicator function $\delta(c_i c_j)$ takes the value of 1 if nodes $i$ and $j$ belong to the same community and 0 otherwise. The range of modularity values is $[-\frac{1}{2}, 1]$, where higher values indicate a better network division into communities. To evaluate the connectivity relationships between clients, we utilize modularity and employ Louvain's algorithm\cite{blondel2008} to obtain the optimal community partitioning of clients. The modularity and optimal community partitioning allow us to assess the connectivity relationships and clustering among clients in the DAG-ACFL.

\begin{figure}
    \centering
    \includegraphics[width=0.5\textwidth]{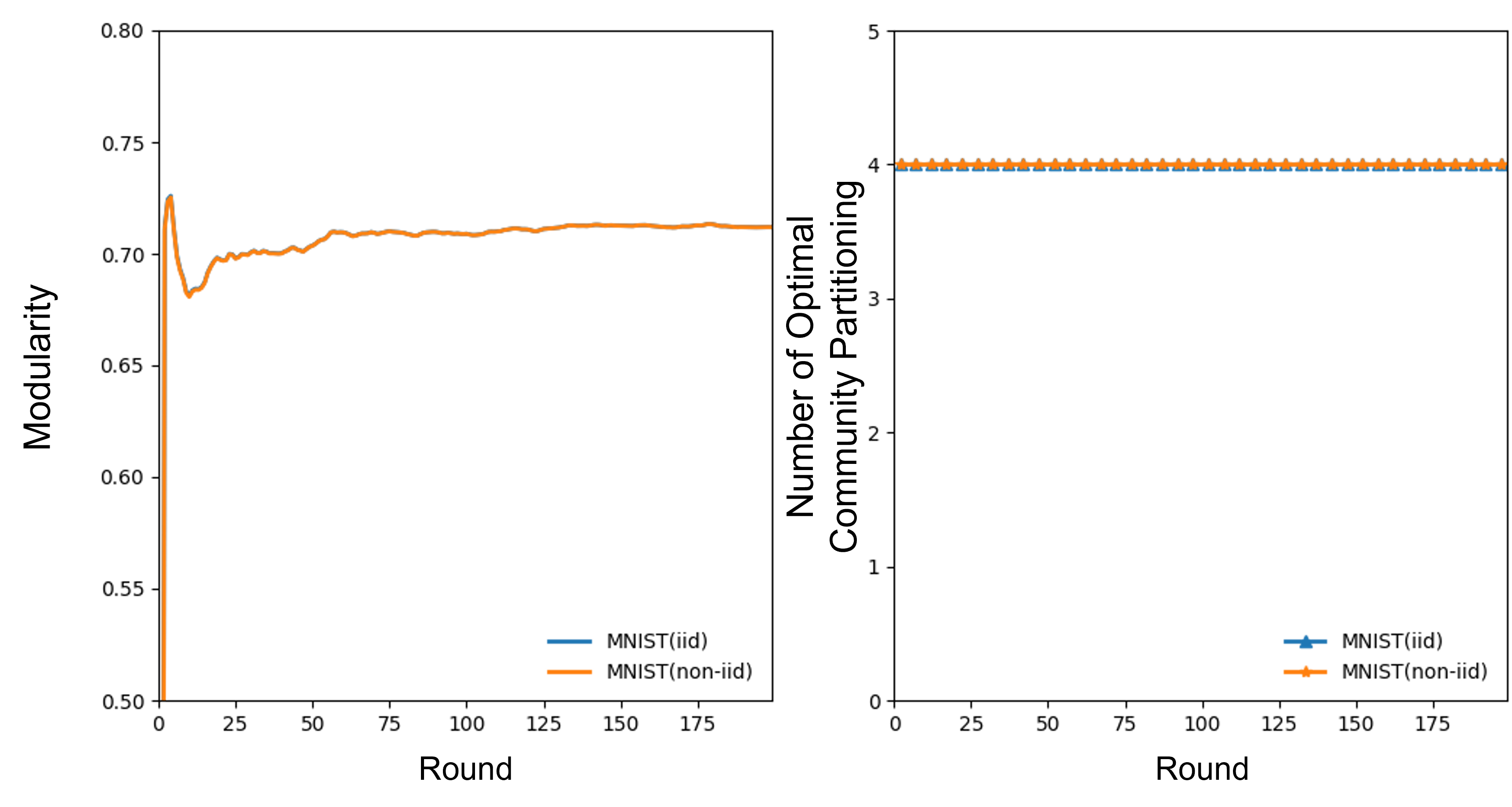}
    \caption{Evaluation results of modularity and optimal community partitioning of clients with CNN model on CIFAR10(type I) and CIFAR10(type II) datasets. The participating rate of the clients is 0.5, and we run the FL process for 200 rounds.}
    \label{fig:modularity_and_modules}
\end{figure}

We evaluate the training process's modularity and optimal community partitioning of clients using the CNN model with CIFAR10(type I) and CIFAR10(type II) datasets. The evaluation result is shown in Figure \ref{fig:modularity_and_modules}. 

From Figure \ref{fig:modularity_and_modules}, we can see that the modularity gradually increases and gradually converges on both the CIFAR10(type I) and CIFAR10(type II) datasets as the training proceeds. It indicates that our method ensures stable connectivity relationships between clients and can effectively cluster them. 
Compared with modularity on the CIFAR10(type I) dataset, the modularity on the CIFAR10(type II) dataset is a little smaller, which dues to the label distribution of the CIFAR10(type II) dataset is more complex.

In addition, we can see that the optimized community partitioning computed by Louvin's algorithm is gradually stabilized as the training proceeds. The number of optimal community partitioning is three on the CIFAR10(type I) and CIFAR10(type II) datasets, consistent with the number of superclasses in the CIFAR10 dataset. This result indicates that our method can effectively cluster the clients into the corresponding superclass. 

\subsection{Evaluation of adaptive tip selection algorithm}
In this section, we evaluate the adaptive tip selection algorithm's performance and use two metrics to do so. The first metric analyzes the satisfaction rate $mr$ of the cluster number corresponding to the tip selected during the training process with the cluster number of all tips from the perspective of a single client, starting with the following computation procedure:

\begin{equation}
	mr= \frac{|selected\_tips_{tip\_cluster\_id==client\_cluster\_id}|}{|all\_tips_{tip\_cluster\_id==client\_cluster\_id}|}
	\label{eq:mr}
\end{equation}	

\begin{figure}
	\centering
	\includegraphics[width=0.5\textwidth]{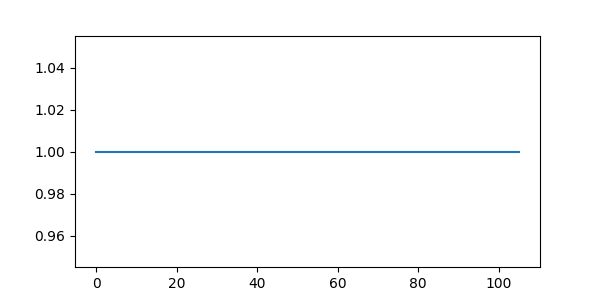}
	\caption{Evaluation of the satisfaction rate $mr$ during the training process and the misclassification during the whole experiment process with 0.5 client participation rate on CIFAR10(type II) dataset. We randomly select another client to evaluate the satisfaction rate $mr$ during the training process and the misclassification during the whole experiment process.}
	\label{fig:mr}
\end{figure}

Another metric is evaluated from an entire client perspective, where we evaluate the performance of our adaptive tip selection algorithm by constructing a neighbor matrix clustering clients and then computing the misclassification of the classifications obtained. We evaluated the most complex experimental case of the above experiments, i.e., using a client participation rate of 0.5 on the CIFAR10 (non-iid) dataset. We randomly chose another client to perform the evaluation period training process with the satisfaction rate $mr$ and the misclassification throughout the experiment. Figure \ref{fig:mr} shows how the satisfaction rate $mr$ of the randomly selected client varies during the training process. We can see that the satisfaction rate is consistently one during the training process, indicating that our method can correctly select all the tip nodes with the same clustering id as the client, and our adaptive tip selection algorithm can efficiently select tips to meet the client's demand.

\begin{figure}
	\centering
	\includegraphics[width=0.5\textwidth]{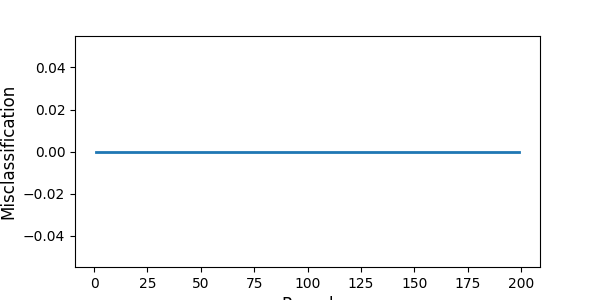}
	\caption{Misclassification per round}
	\label{fig:misclassification}
\end{figure}

Figure \ref{fig:misclassification} shows clients' misclassification during the training process in each round. We can see that the misclassification cases during the training process are always 0, indicating that our adaptive tip selection algorithm can effectively select tips to meet clients' needs.

\subsection{Communication and storage resource Consumption Analysis}
\label{sed:resource}
In this Section, we focus on the communication and storage consumption of DAG-ACFL compared to the baselines for both. To simplify our analysis, we assume that all $|C|$ clients participate in the FL process, and the number of servers in DAG-ACFL is denoted as $|S|$. In our analysis, a transaction is divided into two main parts: the model parameters $\omega$ and the additional information $\sigma$ required to package the transaction. To evaluate the communication consumption, we examine the total amount of data transferred to clients and servers in the one-round FL process. Regarding storage consumption, we consider the storage requirements for storing the model parameters, transaction data, and any other necessary information on both the client and server sides.

{\bf{Communication Consumption Analysis}}

In our analysis of communication consumption in the DAG-ACFL framework, we consider two main parts for clients in each training round: communication consumption before and after local training. The communication consumption can be represented by Equation \ref{eq:DAGACFL-comm}, where the fixed size of the hash value is denoted as $|hash|$.
\begin{equation}
    \begin{aligned}
    \underbrace{|hash| + |\omega|}_{before\ training} + \underbrace{|hash| + |\omega| + 2(|S|-1)\cdot(|\omega| + |\sigma|)}_{after \ training} 
    \end{aligned}
    \label{eq:DAGACFL-comm}
\end{equation}

The communication consumption before local training refers to the data transmission between clients and servers before the local training phase. This process includes the communication consumption of the client sending the $hash$ to the server and the server sending back aggregated model parameters. The communication consumption after local training refers to the data transmission between the server with the client and other servers after the local training phase. This process includes the communication consumption of the client, sending the updated model parameters back to the server, and the server sends the $hash$ value of the transaction to the client and the transaction to other servers. In addition, the server also receives the transaction from other servers.

The communication consumption of the SDAFGL contains two parts: the communication consumption of broadcasting the new transaction created by the client to all other clients and the communication consumption of receiving the transaction from other clients. It is shown in Equation \ref{eq:SDAGFL-comm}.
\begin{equation}
    \begin{aligned}
    \underbrace{(|C|-1)\cdot(|\omega| + |\sigma|)}_{broadcast} + \underbrace{(|C|-1)\cdot(|\omega| + |\sigma|)}_{receive} 
    \end{aligned}
    \label{eq:SDAGFL-comm}
\end{equation}

For FedAvg, FlexCFL, IFCA, and FeSEM, the communication consumption contains two parts, receiving the global model and uploading the model update. Therefore the total communication consumption  of different frameworks for one round of training is shown in Table \ref{tab:communication}
\begin{table}[h]
    \centering
    \caption{Comparison of Total Communication Consumption for One Round of Training in Different Frameworks}
    \begin{tabular}{lc}
      \toprule
      {\bf{Framework}} & {\bf{Communication Consumption}} \\
      \midrule
      {\bf{Fedavg}} & $2\cdot|C|\cdot|\omega|$  \\
      {\bf{FlexCFL}} & $2\cdot|C|\cdot|\omega|$ \\
      {\bf{IFCA}} & $(1+g)\cdot|C|\cdot|\omega|$\\
      {\bf{FeSEM}}& $2\cdot|C|\cdot|\omega|$ \\
      {\bf{SDAGFL}} & $2\cdot|C|\cdot(|C|-1)\cdot(|\omega| + |\sigma|)$  \\
      {\bf{DAG-ACFL}} & $ 2\cdot|C|\cdot\{|hash|+|\omega| + (|S|-1)\cdot(|\omega|+|\sigma|)\}$\\
      \bottomrule
    \end{tabular}

	\begin{tablenotes}
		\item[a] $g$ is the number of predefine cluster
	\end{tablenotes}
    \label{tab:communication}
\end{table}

The analysis of communication consumption presented in Table \ref{tab:communication} demonstrates that DAG-ACFL achieves a significantly lower communication consumption than SDAFGL. The communication consumption in DAG-ACFL is reduced to $\frac{|S|}{|C|}$ of the communication consumption in SDAFGL, where $|S|$ represents the number of servers and $|C|$ represents the number of clients. Notably, the number of servers ($|S|$) is typically much smaller than that of clients ($|C|$). Furthermore, DAG-ACFL effectively reduces the communication burden on clients by shifting it to the servers. This redistribution of communication responsibilities significantly decreases communication consumption for clients.

When comparing DAG-ACFL with other baselines, it is essential to note that the size of the $|hash|$ value is relatively small and can be considered. Thus, the increase in communication consumption primarily occurs at the server layer, which results from utilizing the DAG-DLT. While there is an increase in communication overhead, it is generally acceptable considering the benefits of increased security and reliability provided by DAG-DLT. Additionally, the server typically possesses sufficient communication bandwidth, which mitigates the impact of increased communication consumption. Therefore, this increased communication consumption is not considered a serious issue.

{\bf{Storage resource Consumption Analysis}}

We define several parameters to analyze the storage resource consumption in DAG-ACFL and other baseline frameworks. The average size of the dataset stored in the client is $|\overline{D}|$. And the depth of the DAG ledger is denoted as $d$, while the average number of transactions in each layer is represented by $m$. Additionally, the size of each node is defined as the sum of the model parameters ($|\omega|$) and the additional information ($|\sigma|$). Therefore, the storage consumption of the DAG ledger is shown in Equation \ref{eq:storage-cons}.
\begin{equation}
    L = d \cdot m \cdot (|\omega| + |\sigma|)
    \label{eq:storage-cons}
\end{equation}

The storage consumption in a single client and server within DAG-ACFL and other baseline frameworks is presented in Table \ref{tab:storage}. 

\begin{table}[h]
    \centering
    \caption{Storage consumption of DAG-ACFL and baselines.}
    \begin{tabular}{lcc}
      \toprule
      {\bf{Framework}} & {\bf{Client}} & {\bf{Server}} \\
      \midrule
      {\bf{Fedavg}} & $|\omega|+|\overline{D}|$ & $|C|\cdot|\omega|$\\
      {\bf{FlexCFL}} & $|\omega|+|\overline{D}|$ & $|C|\cdot|\omega|$\\
      {\bf{IFCA}} & $|\omega|+|\overline{D}|$ & $|C|\cdot|\omega|$\\
      {\bf{FeSEM}} & $|\omega|+|\overline{D}|$ & $|C|\cdot|\omega|$\\
      {\bf{SDAGFL}} & $ L +|\omega| + |\overline{D}|$ & $None$ \\
      {\bf{DAG-ACFL}} &  $|\omega|+|\overline{D}|$ & $L$ \\
      \bottomrule
    \end{tabular}
    \label{tab:storage} 
\end{table}

Table \ref{tab:storage} provides the storage consumption of DAG-ACFL and other baseline frameworks. In DAG-ACFL, the size of the DAG ledger primarily determines the server-side storage consumption, while the model size and dataset size influence the client-side storage consumption. In comparison, SDAGFL places a heavier storage burden on the clients. And, as the ledger size increases in SDAGFL, the storage demand on the clients also escalates. This colossal storage consumption in SDAFGL can be problematic for edge devices with limited storage resources, as it may exceed capacity. Furthermore, in DAG-ACFL, the number of servers is typically much smaller than the number of clients. Consequently, the overall storage resource consumption in DAG-ACFL is significantly lower than that in SDAGFL.

When comparing DAG-ACFL with other frameworks, the clients' storage consumption is generally similar. However, DAG-ACFL does incur increased storage demand on the server. But, this increase is acceptable since servers typically possess ample storage resources.

In summary, DAG-ACFL offers significant advantages in terms of saving communication and storage resources compared to SDAGFL, which makes our framework particularly well-suited for edge devices with limited communication and storage capabilities. By efficiently distributing the communication and storage burden, DAG-ACFL optimizes resource utilization and enhances scalability. Compared to other baseline frameworks, our approach does lead to increased communication and storage resource consumption on the server side. However, this increase is deemed acceptable due to the typically higher communication bandwidth and more abundant server storage resources. Moreover, our framework brings additional benefits, such as improved security and reliability, by utilizing DAG-DLT.DAG-ACFL balances resource consumption, scalability, and security, making it a viable and efficient choice for asynchronous clustered FL scenarios.

\section{Conclusion and Future Work}
\label{sec:conclusion}
In this paper, we propose a novel federated learning framework called DAG-ACFL, which leverages Directed Acyclic Graph (DAG)-based distributed ledger technology (DLT) and an adaptive tip selection algorithm to achieve asynchronous clustered federated learning inspired by the advantages of clustered federated learning in addressing statistical heterogeneity and DAG-based DLT in enabling decentralized asynchronous federated learning. DAG-ACFL can effectively select tips published by clients from the same cluster, enabling implicit client clustering without the need to predefine the number of clusters. Specifically, we design a two-layer asynchronous clustered federated learning framework that realizes client-oriented federated learning and provides detailed framework design and implementation. We design a tip selection algorithm based on the cosine similarity of model parameters to effectively select tips published by clients with similar data distributions. Furthermore, we optimize the algorithm and develop an adaptive tip selection algorithm that dynamically determines the number of selected tips using a change point detection algorithm called BEAST.

We construct cluster-wise non-iid datasets with different distributions among clients within the cluster on the MNIST and CIFAR-10 datasets. We also evaluate the performance of DAG-ACFL under different client participation rate scenarios. Experimental results demonstrate that DAG-ACFL can accurately identify tips published by clients from the same cluster in complex environments and achieves performance comparable to state-of-the-art other clustered FL approaches. Additionally, we analyze the communication and storage resource consumption of DAG-ACFL and compare it with baselines, highlighting the superiority of DAG-ACFL in achieving asynchronous clustered federated learning.

Future research directions include exploring effective incentives for client participation in training within the DAG-ACFL framework and superiority in more complex datasets. Furthermore, deploying our framework in real-world scenarios to validate its effectiveness in practical applications is also a future research direction.

% \section*{Acknowledgment}
% This research was supported by the National Natural Science Foundation of China (Grant Number: 62071151, 61301099)

\bibliographystyle{IEEEtran}
\bibliography{references}

\end{document}